\documentclass[letterpaper, 10 pt, conference]{ieeeconf}  
\usepackage{graphicx}
\usepackage{amsmath}
\usepackage{tikz}
\usepackage{textcomp}
\usepackage{hyperref}
\usepackage{lipsum}
\usepackage{atbegshi}

\IEEEoverridecommandlockouts                              

\title{\LARGE \bf
Hybrid Imitation-Learning Motion Planner for Urban Driving
}

\author{Cristian Gariboldi$^{1}$, Matteo Corno$^{2}$ and Beng Jin$^{3}$
\thanks{$^{1}$Cristian Gariboldi is with the Dipartimento di Elettronica, Informazione e Bioingegneria, Politecnico di Milano, Piazza L. da Vinci, 32, 20133, Milano, Italy and Computer Science and Technology, Xi’an Jiaotong University, Bei Lin Qu, 710049, Xi’an, China.
        {\tt\small gariboldicristian@gmail.com}}%
\thanks{$^{2}$Matteo Corno is Full Professor with the Dipartimento di Elettronica, Informazione e Bioingegneria, Politecnico di Milano, Piazza L. da Vinci, 32, 20133, Milano, Italy.}%
\thanks{$^{3}$Beng Jin is Head of Autonomous Driving Department in Pix Moving, Chuangbai Street, Baiyun District, Guiyang, China.}%
}

\newcommand\copyrighttext{%
  \footnotesize \textcopyright 2024 IEEE. Personal use of this material is permitted.
  Permission from IEEE must be obtained for all other uses, in any current or future
  media, including reprinting/republishing this material for advertising or promotional
  purposes, creating new collective works, for resale or redistribution to servers or
  lists, or reuse of any copyrighted component of this work in other works.
  DOI: 10.1109/ITSC58415.2024.10919508}
\newcommand\copyrightnotice{%
\begin{tikzpicture}[remember picture,overlay]
\node[anchor=south,yshift=10pt] at (current page.south) {\fbox{\parbox{\dimexpr\textwidth-\fboxsep-\fboxrule\relax}{\copyrighttext}}};
\end{tikzpicture}%
}

\begin{document}

\maketitle

\thispagestyle{empty}

\pagestyle{empty}

\copyrightnotice
\begin{abstract}

With the release of open source datasets such as nuPlan and Argoverse, the research around learning-based planners has spread a lot in the last years. Existing systems have shown excellent capabilities in imitating the human driver behaviour, but they struggle to guarantee safe closed-loop driving. Conversely, optimization-based planners offer greater security in short-term planning scenarios. To confront this challenge, in this paper we propose a novel hybrid motion planner that integrates both learning-based and optimization-based techniques. Initially, a multilayer perceptron (MLP) generates a human-like trajectory, which is then refined by an optimization-based component. This component not only minimizes tracking errors but also computes a trajectory that is both kinematically feasible and collision-free with obstacles and road boundaries. Our model effectively balances safety and human-likeness, mitigating the trade-off inherent in these objectives. We validate our approach through simulation experiments and further demonstrate its efficacy by deploying it in real-world self-driving vehicles.

\end{abstract}

\section{INTRODUCTION}

Autonomous cars are expected to play a crucial role in future mobility due to their potential for increased safety and road utilization. To ensure these benefits, their planning components must provide safe, comfortable, and collision-free trajectories that account for both static and dynamic traffic elements. Traditional trajectory planning approaches include rule-based, sample-based, and optimization-based methods, which rely on manually defined costs and objective functions optimized using classical techniques like A*, RRT, dynamic programming, and Model Predictive Trajectory algorithms. These methods are reliable and interpretable but struggle to scale in complex urban scenarios and do not improve with data, requiring extensive engineering effort for tuning.

The availability of open-source datasets such as nuPlan and Argoverse has advanced research in learning-based planners, which are very good at generating human-like trajectories. However, these models trained in open-loop settings do not guarantee safety in closed-loop applications, especially in novel scenarios, due to their dependence on training data. To address these limitations, perturbations can be introduced into training datasets to help vehicles recover from dangerous situations and mitigate covariate shift problems. Alternatively, a differentiable simulator can be used for closed-loop training. Despite these improvements, learning-based models still struggle to generalize well in unseen domains, making them unsafe for real-world traffic.

The paper proposes two key contributions:

1) Integration of learning-based and optimization-based techniques to create a hybrid imitation-learning model. This combination aims to generate safe, human-like trajectories, balancing the trade-offs between these objectives. This approach is the first of its kind.

2) Validation of the hybrid model on a real vehicle in urban environments, demonstrating its practical effectiveness and robustness beyond simulation.

Most research in this field is confined to simulations, which may not translate to real-world performance. The goal is to improve the short-term planning capabilities of learning-based models, ensuring their safety and reliability in real urban settings. The research focuses on planning, assuming that localization, perception, mapping, and control modules are already in place.

\section{RELATED WORK}

Generating a comfortable, feasible and collision-free trajectory is a complex task for autonomous driving that has attracted considerable academic interest with several approaches proposed.

\subsection{Optimization-based planners} 
Rule-based and sample-based approaches have been valuable for global and local trajectory planning [1, 2]. However, their complexity makes them unsuitable for real-world autonomous driving in complex scenarios. Consequently, optimization-based planners [3-9] have been proposed, which find optimal trajectories by minimizing predefined cost functions and apply the best control actions for tracking.

Despite their advantages, optimization-based planners face significant challenges:

1) They often struggle to find the global optimum in complex scenarios, as real-time solutions to these optimization problems are difficult, frequently resulting in convergence to local minima in non-convex problems.

2) Even when these planners generate safe, collision-free trajectories, the paths differ significantly from those a human would choose. This discrepancy can confuse and destabilize other agents around the self-driving vehicle, who are not used to predicting the behavior of autonomous cars, potentially leading to unsafe situations.

To address these issues, researchers have turned to machine learning approaches, which have shown promise due to recent advancements in the field.

\subsection{Reinforcement Learning}
Reinforcement Learning for autonomous driving [10-13] removes some human engineering complexity since it uses Machine Learning techniques to learn an optimal policy by maximizing a reward (cost) function, by exploring and exploiting the environment. Even though it is possible to obtain good performance in simulation or
in laboratory's experiments, its performance does not easily translate to real world and complex scenarios and cannot guarantee safety in every driving conditions. This may happen because of its difficulty
to converge to stability.

\subsection{Imitation-Learning}
Imitation-learning models learn driving policies from expert demonstrations, mapping states to actions. Recently, large datasets of human driving behavior have been released by companies and open-source projects such as Argoverse \cite{c15}, Lyft \cite{c16}, Waymo \cite{c17}, and nuPlan \cite{c18}, enhancing the development of imitation-learning in autonomous driving. This approach has led to state-of-the-art solutions for motion forecasting \cite{c19} and robust path planning capabilities.

Our work focuses on leveraging imitation-learning for motion planning by analyzing several methods in this category. ChaufferNet \cite{c20} uses a convolutional neural network to encode a top-down representation of the environment, training it to imitate human driving. The Urban Driver model \cite{c21} optimizes trajectories using a policy gradient method and a differentiable simulator for closed-loop training. In contrast, the Neural Motion Planner system \cite{c14} uses sensor and HD map data to generate 3D detections, future trajectories, and a cost volume, selecting the trajectory with the minimum learned cost.

A multimodal prediction strategy combines a transformer with a Mixture of Experts approach \cite{c22} to model probability distributions over multiple future trajectories, selecting the one minimizing a predefined cost function. Hybrid models, like SafetyNet \cite{c23}, integrate a machine learning planner with a rule-based fallback layer to ensure trajectory feasibility and safety, executing either the ML or fallback trajectory based on dynamic feasibility checks.

Another hybrid model, PDM-Hybrid \cite{c24, c25}, uses trajectory fusion between sample-based and learning-based planners to achieve high scores in the nuPlan simulator. However, this model presents several issues:

1) The fusion of the two trajectories involves linear interpolation based on a correction horizon, denoted as C. Up to C, the trajectory is guided by the sample-based approach, transitioning to the learning-based trajectory beyond C. However, this method may introduce discontinuities in the final trajectory due to inconsistencies at the fusion point C;

2) While this strategy aims to produce a prediction trajectory resembling human behavior (after C), the actual path taken by the ego-vehicle aligns with the output of the sample-based approach. Consequently, the final trajectory may lack human-like characteristics, deviating from expected human behavior.

\section{SYSTEM ARCHITECTURE}
This section describes our hybrid imitation-learning model, combining a learning-based planner with an optimization-based component for kinematically feasible, collision-free trajectories. As outlined in Fig. 1, the system inputs the ego vehicle states, perception observations, and a goal destination to generate a sample-based trajectory with the Planner block. A Multilayer Perceptron (MLP) refines this trajectory to mimic human-like behavior, and the Model Predictive Trajectory (MPT) block optimizes it to avoid collisions with obstacles and road boundaries.

\begin{figure}[thpb]
    \centering
    \framebox{\parbox{3.3in}{\includegraphics[width=\linewidth]{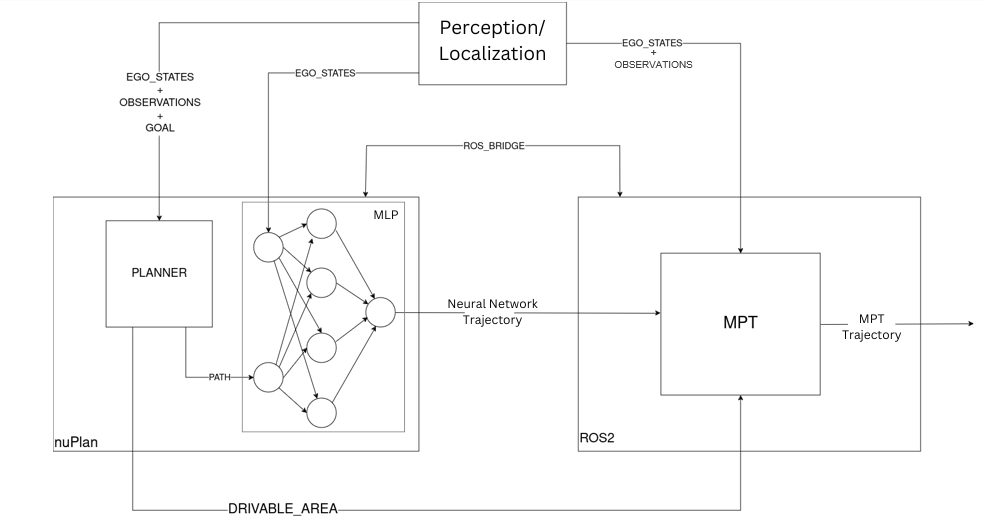}}}
    \caption{Model's Structure | This is an overview of the hybrid model, showing the inputs and outputs of each block and the communication between interfaces. We also show the inputs provided by the Perception and Localization components such as observations \cite{c26}, ego states and goal destination.}
    \label{Model's Structure | Training Stage}
\end{figure}

\subsection{Planner}
The planner block together with the Multilayer Perceptron, was inspired by PDM-Open model [24, 25], which,
taking as inputs the poses, velocities and accelerations of the ego vehicle, the observations (used for agents forecasting) and the goal, it is responsible to find a centerline from the starting position to the end point, leveraging on the Dijkstra algorithm \cite{c27}, and to compute a collision-free path,
relying on a sample-based approach.

The planner computes 15 different paths in the following way:

1) Starting from the centerline, it employs 5 different Intelligent Driver Model (IDM) \cite{c28} policies with specific target speeds, specifically 20\%, 40\%, 60\%, 80\% and 100\% of the speed limit. When there is a leading vehicle in front of the ego, the speed limit is defined as the velocity of the leading vehicle;

2) Secondly, in order to have lateral variance, we also apply 3 different offsets from the centerline, respectively +1m, -1m and 0m.

This way, we have 15 different paths with longitudinal and lateral variety which are simulated in the forecasted environment and scored according to the closed-loop metrics provided by nuPlan.
The path with the highest score is then selected and if it has an expected at-fault collision within 2 seconds, the output is overwritten with a maximum braking force maneuver.

\subsection{Multilayer Perceptron (MLP)}
The multilayer perceptron is responsible for generating an
output trajectory which should be as similar as possible to the expert
driver one. In order to achieve this task, the neural network takes as inputs the ego vehicle's poses, velocities and accelerations of
longitudinal, lateral and angular axis, starting from the past 2 seconds up to the current
time step, together with the path computed by the planner block. These inputs are scaled to a 512-dimensional vector using a linear layer and then they are concatenated and fed into the MLP.

The MLP consists of two 512-dimensional linear layers with dropout (p=0.1) and ReLU activation functions. The output layer is a linear layer that regresses the future waypoints for the next 8 seconds. This output is called "Neural Network Trajectory," trained to minimize the L2 distance between its waypoints and those of the expert driver trajectory provided by the dataset, which offers more than 88-thousands scenarios with a length of 15 seconds with human driver trajectories for training purposes.

\subsection{Model Predictive Trajectory (MPT)}
The optimization-based component utilizes a MPT algorithm. This algorithm integrates inputs such as the "Neural Network Trajectory" generated by the MLP, the drivable area, the ego vehicle's poses and velocities and the observations from the perception system. Its primary function is to produce an optimized trajectory that ensures both collision-free navigation and adherence to kinematic feasibility.

Aiming to solve an optimization problem, we define the following soft and hard constraints:

1) \textbf{Soft Constraint:} the collision-free condition is considered as a soft constraint, since if the optimized trajectory is not collision-free, we take into consideration the previously generated trajectory;

2) \textbf{Hard Constraint:} since the trajectory near the ego vehicle must be smooth, the only hard constraint we have is that the trajectory points near the ego must be the same as the previously generated trajectory, in order to avoid sudden steering maneuvers. This hard constraint is formulated as follow:
    \[\delta_k = \delta_k^{prev} \quad \textrm{if} \quad (0 \le i \le N_{fix})\]

    Where:
    \begin{itemize}
        \item \(\delta_k\) represents the steering angle at a current trajectory point;
        \item \(\delta_k^{prev}\)  represents the steering angle at the previous trajectory point. It ensures that the current steering angle remains consistent with the previous one;
        \item \(N_{fix}\) represents the number of fixed trajectory points. It determines the range over which the hard constraint is applied.
    \end{itemize}

The objective function of the optimization problem minimizes the tracking errors and the steering acceleration, rate and angle of the ego vehicle. 

It can be defined as follow:


\begin{equation}
\begin{aligned}
    J &=  w_y \sum_{k} y_k^2 + w_{\theta} \sum_{k} \theta_k^2 + w_{\delta} \sum_{k} \delta_k^2 \\
    & + w_{\dot{\delta}} \sum_{k} \dot{\delta_k^2} + w_{\ddot{\delta}} \sum_{k} \ddot{\delta_k^2} \\
\end{aligned}
\end{equation}

Where at time step \(k\), we can define the following variables:
\begin{itemize}
    \item \(y_k\): lateral distance to reference path;
    \item \(\theta_k\): heading angle against the reference path;
    \item \(\delta_k\): steering angle;
    \item \(\dot{\delta_k}\): steering rate;
    \item \(\ddot{\delta_k}\): steering acceleration.
    \item \(w_y, w_{\theta}, w_{\delta}, w_{\dot{\delta}}, w_{\ddot{\delta}}\) are tuning weights.
\end{itemize}

The MPT, by taking as input the observations of other agents, is also able to perform adaptive cruise planning maneuvers. The role of the cruise planning is keeping a safe distance with dynamic vehicle objects with smoothed velocity transition.

The safe distance is calculated dynamically by the following equation:

\[d = v_{ego} t_{idling} + \frac{1}{2} a_{ego} t_{idling}^2 + \frac{v_{ego}^2}{2a_{ego}} - \frac{v_{obstacle}^2}{2a_{obstacle}}\]

where:

\begin{itemize}
    \item \(d\) is the calculated safe distance;
    \item \(t_{idling}\) is the idling time for the ego to detect the front vehicle's deceleration;
    \item \(v_{ego}\) is the ego's current velocity;
    \item \(v_{obstacle}\) is the front obstacle's current velocity;
    \item \(a_{ego}\) is the ego's acceleration;
    \item \(a_{obstacle}\) is the obstacle's acceleration.
\end{itemize}

To maintain a safe distance while optimizing for smooth velocity transitions, we solve an optimization problem. The objective function minimizes the deviation from the desired velocity and smoothness of acceleration:

\begin{equation*}
    J =  \sum_{k} (w_v(v_{desired} - v_{ego, k})^2 + w_a a_{ego, k}^2)
\end{equation*}

subject to constraints on safe distance \(d\), velocity, and acceleration. By solving this problem at each time step, the ego vehicle adapts to changes and ensures safe and efficient cruising. (Note that \(w_v, w_a\) are tuning weights).







\section{EXPERIMENTS AND RESULTS}
\subsection{Baselines}
We first analyze the results from the nuPlan open-loop (OL), closed-loop non-reactive (CL-NR), and closed-loop reactive (CL-R) simulations for baseline models, as shown in TABLE 1. Scores are computed using the simulator's built-in metrics. Open-loop simulations evaluate the planner's imitation of an expert driver's route, while closed-loop simulations assess the trajectory's safety, comfort, and collision avoidance. Each simulation assigns a score between 0 and 100 based on these criteria.

\begin{table}[h]
\caption{Baselines' scores in nuPlan}
\label{nuPlan scores}
\begin{center}
\begin{tabular}{|c||c||c||c|}
\hline
Planner & OL & CL-NR & CL-R\\
\hline
Urban Driver & \textbf{82} & 53 & 50\\
\hline
GC-PGP & \textbf{82} & 57 & 54\\
\hline
PDM-Open & \textbf{86} & 50 & 54\\
\hline
IDM & 38 & \textbf{76} & \textbf{77}\\
\hline
PDM-Closed & 42 & \textbf{93} & \textbf{92}\\
\hline
\end{tabular}
\end{center}
\end{table}

Upon closer examination of TABLE 1, a discernible pattern emerges within the results.

Specifically, Urban Driver, PDM-Open and GC-PGP \cite{c29}, characterized as learning-based models, exhibit
commendable performance in open-loop simulations but display diminished efficacy in
closed-loop scenarios.

Conversely, the rule-based IDM and the sample-based PDM-Closed models demonstrate an inverse behaviour: underperforming in
open-loop simulations yet surpassing the learning-based models in closed-loop simulations.

These findings suggest that learning-based models excel in predicting the motion of the ego
vehicle, capable of replicating human trajectories. However, unlike rule, sample or optimization-based
approaches, they do not inherently ensure safe closed-loop driving.

\subsection{ROS Simulator}
Before testing the model directly on the real vehicle, several experiments have been conducted in the simulator.

Fig. 2 shows different experimental results.

\begin{figure}[thpb]
    \centering
    \framebox{\parbox{3.3in}{\includegraphics[width=\linewidth]{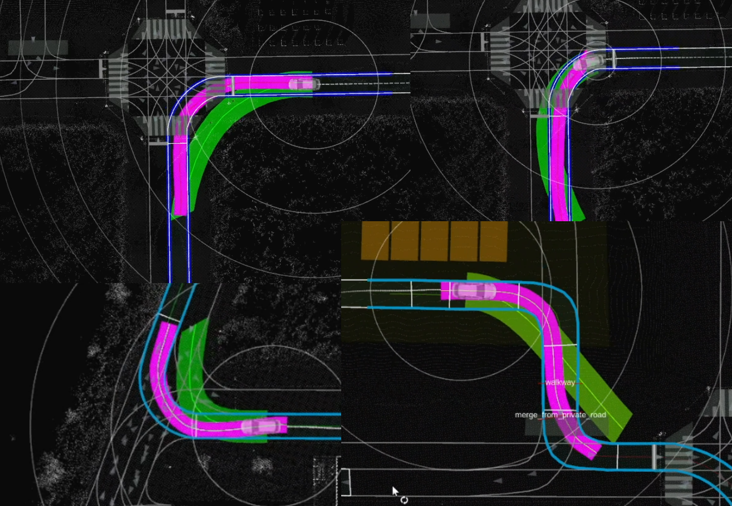}}}
    \caption{Four frames from self-driving simulations, the green line shows the neural network's output ("Neural Network Trajectory") and the pink line shows the optimization-based component's output ("MPT Trajectory"). The optimization process adjusts the neural network's output to avoid collisions with obstacles and road boundaries.}
    \label{Experimental Results}
\end{figure}

The green line is the "Neural Network Trajectory", direct output of the neural network. As expected it is not able to provide a safe closed-loop driving, as in the provided corner cases of Fig. 2 it often overcomes the boundaries of the lane, leading to unsafe and dangerous situations without the guarantee of a collision-free trajectory. Despite that, it shows good generalization capabilities, as the maps and scenarios considered during evaluation are completely different from the ones in the training stage. 

However, the pink line, which represents the "MPT Trajectory", perfectly drives the vehicle within the bounds of the lane, redefining the multilayer perceptron's output into a safe and collision-free route.

The model is also able to perform collision avoidance maneuvers with static obstacles and adaptive cruise control driving with dynamic agents.

Thanks to these experimental results, it is possible to demonstrate the effectiveness of the safe closed-loop driving capabilities of the hybrid motion planner, which is indeed able to prevent collisions and unfeasible trajectories by computing a refined output through the optimization process.

However, assessing the model's ability to mimic human-like driving style requires a qualitative analysis.

To this aim, we examine several qualitative results.

The following results in Fig. 3-8 show some comparisons between a default optimization-based planner (on the left), and the hybrid motion planner that we propose in this paper (on the right).

In addition to the shape of the trajectories, also the velocity (top) and acceleration (bottom) profiles are provided in order to better evaluate the human-likeness.

In Fig. 3, the trajectories of both the default and hybrid planners exhibit a striking similarity in shape. However, upon closer inspection, we notice an interesting distinction: the hybrid model's trajectory gracefully widens around curves, diverging from the lane centerline, mirroring human driver behavior more closely.

\begin{figure}[thpb]
    \centering
    \framebox{\parbox{3.3in}{\includegraphics[width=\linewidth]{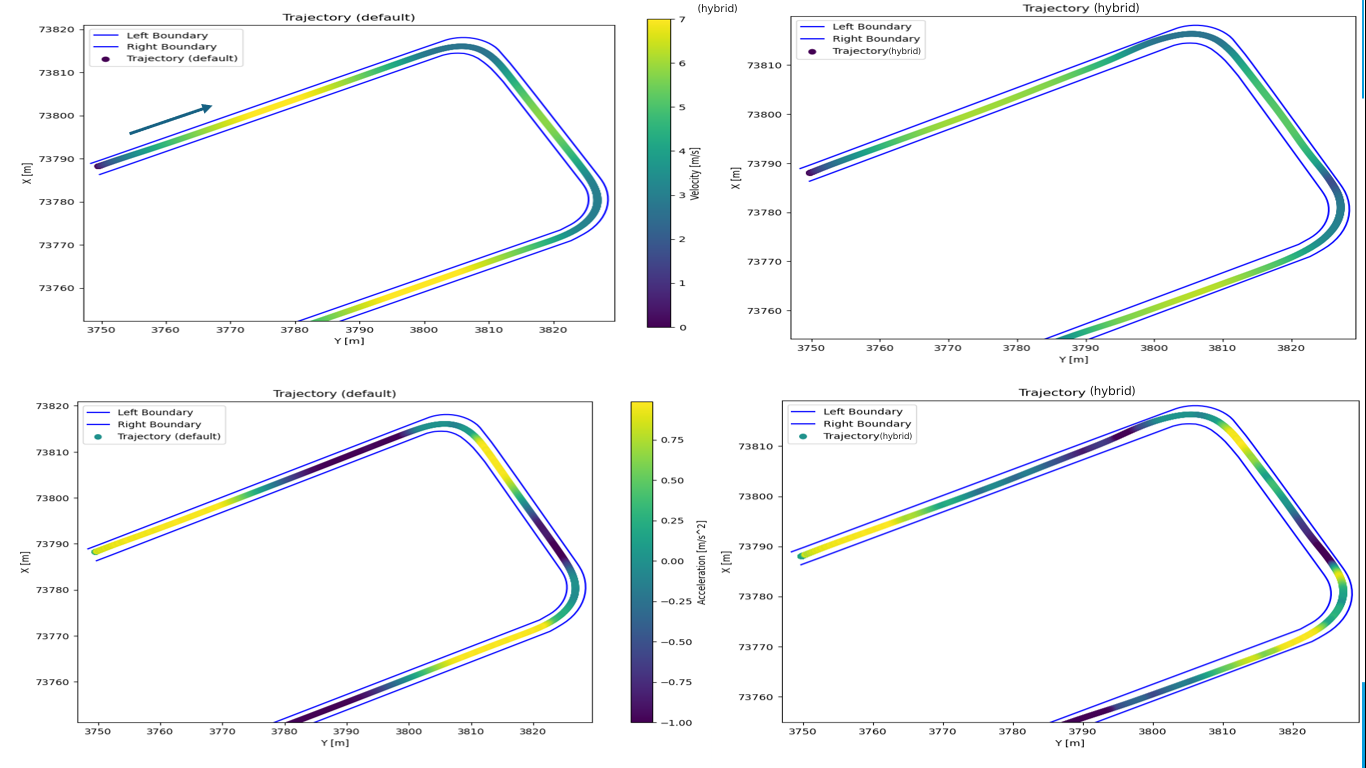}}}
    \caption{Left: default optimization-based planner's trajectory with velocity (top) and acceleration (bottom). Right: hybrid model trajectory with velocity (top) and acceleration (bottom). The vehicle's motion follows the arrow in the top left image. Blue lines indicate the lane boundaries.}
    \label{Qualitative Result 2}
\end{figure}

Furthermore, the velocity and acceleration profiles of the hybrid planner are considerably smoother compared to the optimization-based model. In the latter, abrupt maneuvers and accelerations are evident, resulting in a discontinuous overall motion.


In Fig. 4, although the velocity and acceleration profiles looks very similar among the two planners, we can distinguish a notable difference in the shape of the trajectories. While the default planner almost perfectly follows the lane centerline, leading to a geometrical path, the hybrid model moves away from the centerline, driving along the two turns with a single maneuver.

\begin{figure}[thpb]
    \centering
    \framebox{\parbox{3.3in}{\includegraphics[width=\linewidth]{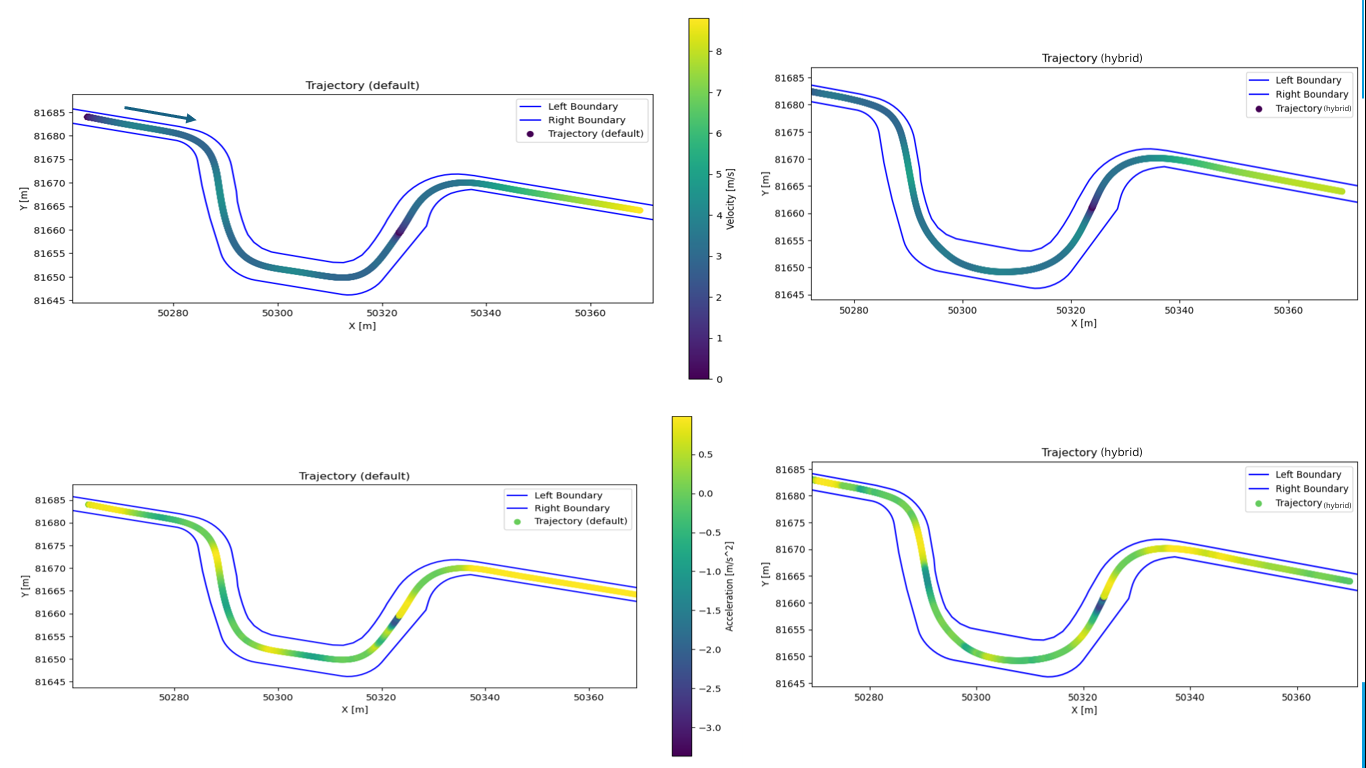}}}
    \caption{Same as Fig. 3}
    \label{Qualitative Result 4}
\end{figure}

Similarly, in Fig. 5 we obtain a human-like trajectory with the hybrid model, which widens around curves. Moreover, on the top of the image, we can notice an interesting behaviour in the shape of the trajectories. In that spot, we can see there is an abrupt step in the right bound of the lane, which also affects the default planner trajectory. Conversely, the hybrid planner completely ignore the step in the lane bound and does not affect the motion, leading to a more comfortable route.

\begin{figure}[thpb]
    \centering
    \framebox{\parbox{3.3in}{\includegraphics[width=\linewidth]{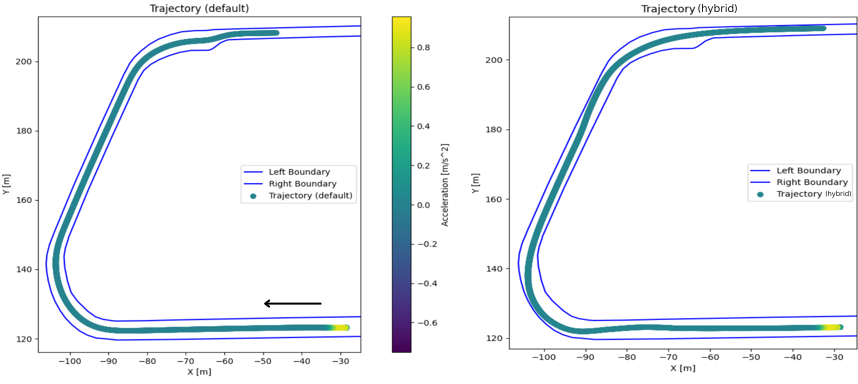}}}
    \caption{Same as Fig. 3}
    \label{Qualitative Result last}
\end{figure}

Another interesting case is shown in Fig. 6, where an adaptive cruise control maneuver was simulated.

While the default planner abruptly accelerates at the beginning, reaching high velocities in a short time, and suddenly brakes when encountering the leading vehicle, leading to an uncomfortable and discontinuous motion, the hybrid planner employs a much smoother trajectory, inferring the right acceleration to avoid brusque and curt maneuvers.

As the experiment was conducted along a straight line, with no discernible differences between the two trajectories, we shift our focus to analyzing velocity and acceleration in the time domain in Fig. 7, where we can notice much smoother profiles in the hybrid model's motion.

\begin{figure}[thpb]
    \centering
    \framebox{\parbox{3.3in}{\includegraphics[width=\linewidth]{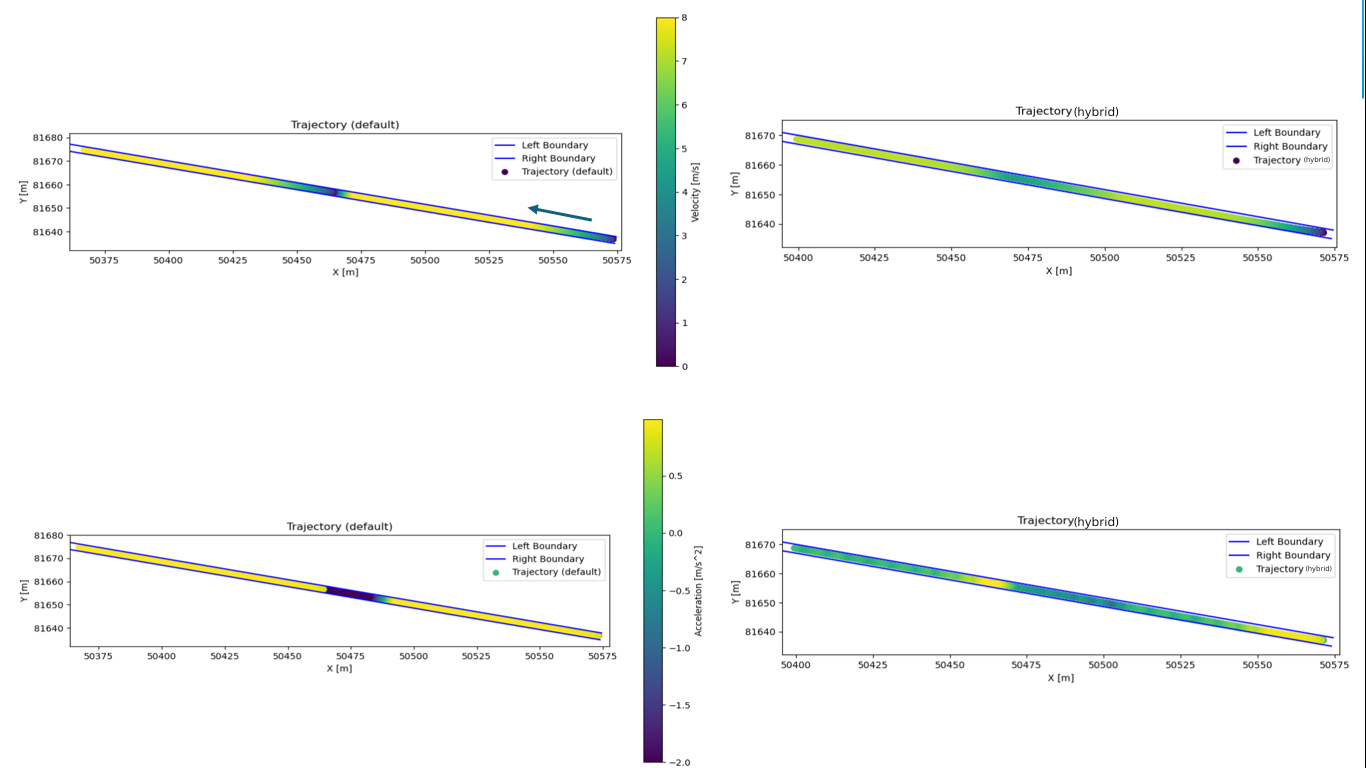}}}
    \caption{Same as Fig. 3}
    \label{Qualitative Result 5}
\end{figure}

\begin{figure}[thpb]
    \centering
    \framebox{\parbox{3.3in}{\includegraphics[width=\linewidth]{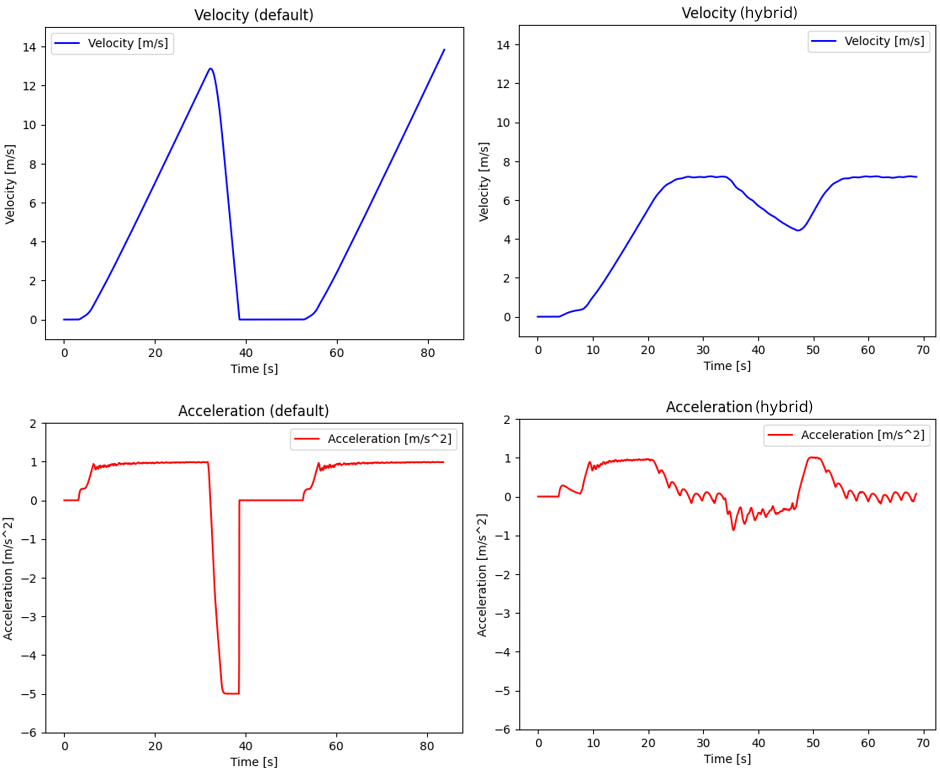}}}
    \caption{Results of velocity and acceleration profiles in time domain during an adaptive cruise control maneuver. Left: default optimization-based planner. Right: hybrid planner.}
    \label{Velocities and Accelerations in Time Domain}
\end{figure}

In addition to our qualitative analysis, we shift our focus to quantitative results by inspecting the jerk profile in the time domain. Elevated jerk levels are characteristic of robotic maneuvers, whereas moderated levels reflect a more human-like driving style.

In Fig. 8, we can notice a high peak in the jerk profile of the optimization-based planner, while the hybrid motion planner's one remains contained.

\begin{figure}[thpb]
    \centering
    \framebox{\parbox{3.3in}{\includegraphics[width=\linewidth]{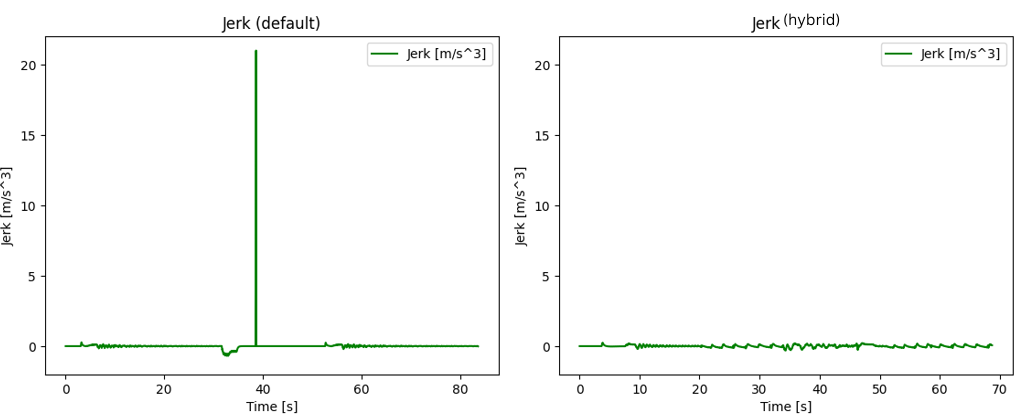}}}
    \caption{Jerks in Time Domain during an adaptive cruise control maneuver. Left: default optimization-based planner. Right: hybrid model.}
    \label{Jerks in Time Domain}
\end{figure}

\subsection{Real World Driving}
After having conducted several successful tests in the simulator environment, we now shift our case study into real-world scenarios. The model has been deployed on the vehicle built and designed by the company Pix Moving, which is called Robobus. The Robobus is a bi-directional, level 4 autonomous vehicle, fully electric, with sensors such as lidars, radars, cameras, GNSS and IMU. It has been designed to transport up to six people, with a maximum speed of 30km/h and it is already operating in some areas in China and Japan. The experiments took place in a real traffic scenarios, with other static and dynamic agents involved, as shown in Fig. 9. The planner showed stable and robust performance while navigating in the traffic, especially at low speed (less than 15km/h). Thanks to the optimization-based component which refines the output of the neural network, the final trajectory was always within the lane boundaries and collision-free with obstacles and other agents.


\section{CONCLUSIONS}

In our paper, we introduce a hybrid imitation-learning motion planner designed to ensure safe, collision-free trajectories that closely mimic human-like behavior. Our model exhibits impressive performance in simulation, demonstrating strong generalization across diverse maps, scenarios, and environments not seen during training. This underscores its robust capabilities. Moreover, our approach proves effective when deployed in real-world self-driving vehicles, particularly at low speeds. As we move forward, future research efforts should prioritize testing the model at higher speeds to better prepare it for real-world urban driving scenarios.

\begin{figure}[thpb]
    \centering
    \framebox{\parbox{3.3in}{\includegraphics[width=\linewidth]{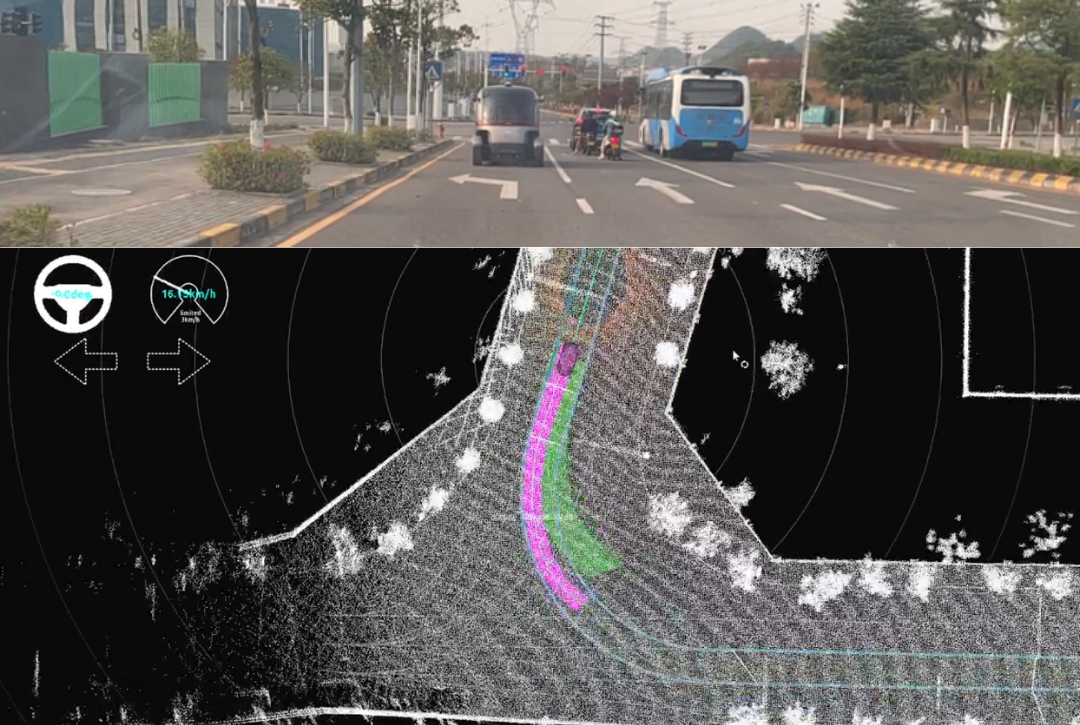}}}
    \caption{Real-world Test. Top: front view from the ego vehicle in urban traffic. Bottom: top-down sensor visualization with planned trajectories. Green indicates the neural network output, and pink represents the hybrid model's trajectory.}
    \label{Real test}
\end{figure}

\addtolength{\textheight}{-12cm}   





\end{document}